\documentclass[11pt,letterpaper]{article}
\usepackage{emnlp2017}
\pdfoutput=1
\usepackage{times}
\usepackage{latexsym}

\usepackage{mylingmacros}
\usepackage{amsmath}

\emnlpfinalcopy

\def\emnlppaperid{842}

\newcommand\task{{\sc S}plit-and-{\sc R}ephrase}
\newcommand\webnlg{{\sc WebNLG}}
\newcommand\tasking{splitting-and-rephrasing}
\newcommand\websplit{\textsc{WebSplit}}
\def\nl#1{{``#1''}}
\def\nlt#1{{#1}}
\newcommand{\sem}[1]{{\textit{#1}}}

\newcommand\seqseq{\textsc{Seq2Seq}}
\newcommand\mseqseq{\textsc{MultiSeq2Seq}}
\newcommand\hybridsimpl{\textsc{HybridSimpl}}
\newcommand\splitseqseq{\textsc{Split-Seq2Seq}}
\newcommand\splitmseqseq{\textsc{Split-MultiSeq2Seq}}

\title{Split and Rephrase
}

\author{Shashi Narayan$^{\dagger}$ \quad Claire Gardent$^{\ddagger}$
  \quad Shay B. Cohen$^{\dagger}$ \quad Anastasia
  Shimorina$^{\ddagger}$ \\ $^{\dagger}$ \normalsize{School of
    Informatics, University of Edinburgh, 10 Crichton Street,
    Edinburgh EH8 9AB, UK} \\ $^{\ddagger}$ \normalsize{CNRS, LORIA,
    UMR 7503, Vandoeuvre-l\`{e}s-Nancy, F-54500, France}
  \\ \tt{shashi.narayan@ed.ac.uk} \tt{claire.gardent@loria.fr} \\
  \tt{scohen@inf.ed.ac.uk} \tt{anastasia.shimorina@loria.fr}}

\date{}

\begin{document}

\maketitle

\begin{abstract}
We propose a new sentence simplification task (\task ) where the aim
is to split a complex sentence into a meaning preserving sequence of
shorter sentences.  Like sentence simplification, \tasking\ has the
potential of benefiting both natural language processing and societal
applications. Because shorter sentences are generally better processed
by NLP systems, it could be used as a preprocessing step which
facilitates and improves the performance of parsers, semantic role
labelers and machine translation systems. It should also be of use for
people with reading disabilities because it allows the conversion of
longer sentences into shorter ones. This paper makes two contributions
towards this new task. First, we create and make available a benchmark
consisting of 1,066,115 tuples mapping a single complex sentence to a
sequence of sentences expressing the same meaning.\footnote{The
  \task\ dataset is available here:
  \url{https://github.com/shashiongithub/Split-and-Rephrase}.} Second,
we propose five models (vanilla sequence-to-sequence to
semantically-motivated models) to understand the difficulty of the
proposed task.
\end{abstract}

\section{Introduction}

Several sentence rewriting operations have been extensively discussed
in the literature: sentence compression, multi-sentence fusion,
sentence paraphrasing and sentence simplification.

Sentence compression rewrites an input sentence into a shorter
paraphrase
\cite{knight2000statistics,cohn-lapata:2008:PAPERS,filippova2008dependency,pitler2010methods,filippova-emnlp15,Toutanova:emnlp2016}. Sentence
fusion consists of combining two or more sentences with overlapping
information content, preserving common information and deleting
irrelevant details
\cite{mckeown2010time,filippova2010multi,thadani2013supervised}. Sentence
paraphrasing aims to rewrite a sentence while preserving its meaning
\cite{dras1999tree,barzilay2001extracting,bannard2005paraphrasing,wubben2010paraphrase,mallinson:2017:EACL}. Finally,
sentence (or text) simplification aims to produce a text that is
easier to understand
\cite{siddharthan2004syntactic,zhu2010monolingual,woodsend2011learning,wubben2012sentence,narayan2014hybrid,wei15tacl,narayan-inlg16b,zhang-emnlp17}. Because
the vocabulary used, the length of the sentences and the syntactic
structures occurring in a text are all factors known to affect
readability, simplification systems mostly focus on modelling three
main text rewriting operations: simplifying paraphrasing, sentence
splitting and deletion.

We propose a new sentence simplification task, which we dub \task ,
where the goal is to split a complex input sentence into shorter
sentences while preserving meaning. In that task, the emphasis is on
sentence splitting and rephrasing. There is no deletion and no lexical
or phrasal simplification but the systems must learn to split complex
sentences into shorter ones and to make the syntactic
transformations required by the split (e.g., turn a relative clause
into a main clause). Table~\ref{tab:rewriting} summarises the
similarities and differences between the five sentence rewriting
tasks.

\begin{table}[t]
  \centering
\footnotesize{
\begin{tabular}{lcccc}
& Split & Delete & Rephr. & MPre.\\\hline
Compression & N & Y  & ?Y & N\\
Fusion & N & Y  & Y & ?Y\\
Paraphrasing & N & N & Y & Y\\
Simplification & Y & Y  & Y & N\\
\task & Y & N  & Y & Y\\\hline
\end{tabular}}
\caption{Similarities and differences between sentence rewriting tasks
  with respect to splitting (Split), deletion (Delete), rephrasing
  (Rephr.) and meaning preserving (MPre.) operations (Y: yes, N: No,
  ?Y: should do but most existing approaches do
  not).}\label{tab:rewriting}
\end{table}

Like sentence simplification, \tasking\ could benefit both natural
language processing and societal applications.  Because shorter
sentences are generally better processed by NLP systems, it could be
used as a preprocessing step which facilitates and improves the
performance of parsers
\cite{tomita1985efficient,chandrasekar1997automatic,McDonald:coli2011,JELNEK14.228},
semantic role labelers \cite{vickrey2008sentence} and statistical
machine translation (SMT) systems \cite{chandrasekar1996motivations}.
In addition, because it allows the conversion of longer sentences into
shorter ones, it should also be of use for people with reading
disabilities \cite{inui2003text} such as aphasia patients
\cite{carroll1999simplifying}, low-literacy readers
\cite{watanabe2009facilita}, language learners
\cite{siddharthan2002architecture} and children \cite{debelder2010}.

\paragraph{Contributions.} We make two main contributions towards the
development of \task\ systems.

Our first contribution consists in creating and making available a
benchmark for training and testing \task\ systems. This benchmark
(\websplit ) differs from the corpora used to train sentence
paraphrasing, simplification, compression or fusion models in three
main ways.

First, it contains a high number of splits and rephrasings. This is
because (i) each complex sentence is mapped to a rephrasing consisting
of at least two sentences and (ii) as noted above, splitting a
sentence into two usually imposes a syntactic rephrasing (e.g.,
transforming a relative clause or a subordinate into a main clause).

Second, the corpus has a vocabulary of 3,311 word forms for a little
over 1 million training items which reduces sparse data issues and
facilitates learning. This is in stark contrast to the relatively
small size corpora with very large vocabularies used for
simplification (cf. Section \ref{sec:related}).

Third, complex sentences and their rephrasings are systematically
associated with a meaning representation which can be used to guide
learning. This allows for the learning of semantically-informed models
(cf. Section~\ref{sec:baselines}). 


Our second contribution is to provide five models to understand the
difficulty of the proposed \task\ task: (i) A basic encoder-decoder
taking as input only the complex sentence; (ii) A hybrid
probabilistic-SMT model taking as input a deep semantic representation
(Discourse representation structures, \citeauthor{kamp81}
\citeyear{kamp81}) of the complex sentence produced by Boxer
\cite{curran2007linguistically}; (iii) A multi-source encoder-decoder
taking as input both the complex sentence and the corresponding set of
RDF (Resource Description Format) triples; (iv,v) Two
partition-and-generate approaches which first, partition the semantics
(set of RDF triples) of the complex sentence into smaller units and
then generate a text for each RDF subset in that partition. One model
is multi-source and takes the input complex sentence into account when
generating while the other does not.


\section{Related Work}
\label{sec:related}

We briefly review previous work on sentence splitting and rephrasing.

\paragraph{Sentence Splitting.} 
Of the four sentence rewriting tasks (paraphrasing, fusion,
compression and simplification) mentioned above, only sentence
simplification involves sentence splitting. Most simplification
methods learn a statistical model
\cite{zhu2010monolingual,coster2011learning,woodsend2011learning,wubben2012sentence,narayan2014hybrid}
from the parallel dataset of complex-simplified sentences derived by
\newcite{zhu2010monolingual} from Simple English
Wikipedia\footnote{Simple English Wikipedia
  (\url{http://simple.wikipedia.org}) is a corpus of simple texts
  targeting ``children and adults who are learning English Language''
  and whose authors are requested to ``use easy words and short
  sentences''.} and the traditional one\footnote{English Wikipedia
  (\url{http://en.wikipedia.org}).}.

For training \task\ models, this dataset is arguably ill suited as it
consists of 108,016 complex and 114,924 simplified sentences thereby
yielding an average number of simple sentences per complex sentence of
1.06. Indeed, \newcite{narayan2014hybrid} report that only 6.1\% of
the complex sentences are in fact split in the corresponding
simplification.  A more detailed evaluation of the dataset by
\newcite{wei15tacl} further shows that (i) for a large number of
pairs, the simplifications are in fact not simpler than the input
sentence, (ii) automatic alignments resulted in incorrect
complex-simplified pairs and (iii) models trained on this dataset
generalised poorly to other text genres. \newcite{wei15tacl} therefore
propose a new dataset, Newsela, which consists of 1,130 news articles
each rewritten in four different ways to match 5 different levels of
simplicity. By pairing each sentence in that dataset with the
corresponding sentences from simpler levels (and ignoring pairs of
contiguous levels to avoid sentence pairs that are too similar to each
other), it is possible to create a corpus consisting of 96,414
distinct complex and 97,135 simplified sentences. Here 
again however, the proportion of splits is very low.

As we shall see in Section~\ref{subsec:benchmarkresults}, the new dataset
we propose differs from both the Newsela and the Wikipedia
simplification corpus, in that it contains a high number of splits. In
average, this new dataset associates 4.99 simple sentences with each complex
sentence.

\paragraph{Rephrasing.} 
Sentence compression, sentence fusion, sentence paraphrasing and
sentence simplification all involve rephrasing.

Paraphrasing approaches include bootstrapping approaches which start
from slotted templates (e.g.,\nl{X is the author of Y}) and seed
(e.g.,\nl{X = Jack Kerouac, Y = ``On the Road''}) to iteratively learn
new templates from the seeds and new seeds from the new templates
\cite{ravichandran2002learning,duclaye2003learning}; systems which
extract paraphrase patterns from large monolingual corpora and use
them to rewrite an input text \cite{duboue2006answering,narayan-inlg16a}; statistical
machine translation (SMT) based systems which learn paraphrases from
monolingual parallel \cite{barzilay2001extracting,zhao2008pivot},
comparable \cite{quirk2004monolingual} or bilingual parallel
\cite{bannard2005paraphrasing,ganitkevitch:2011:EMNLP} corpora; and a
recent neural machine translation (NMT) based system which learns
paraphrases from bilingual parallel corpora
\cite{mallinson:2017:EACL}.

In sentence simplification approaches, rephrasing is performed either
by a machine translation
\cite{coster2011learning,wubben2012sentence,narayan2014hybrid,wei16tacl,zhang-emnlp17}
or by a probabilistic model
\cite{zhu2010monolingual,woodsend2011learning}.  Other approaches
include symbolic approaches where hand-crafted rules are used e.g., to
split coordinated and subordinated sentences into several, simpler
clauses
\cite{chandrasekar1997automatic,siddharthan2002architecture,canning2002syntactic,siddharthan2010complex,siddharthan2011text}
and lexical rephrasing rules are induced from the Wikipedia
simplification corpus \cite{siddharthan2014hybrid}.

Most sentence compression approaches focus on deleting words (the
words appearing in the compression are words occurring in the input)
and therefore only perform limited paraphrasing. As noted by
\newcite{pitler2010methods} and \newcite{Toutanova:emnlp2016} however,
the ability to paraphrase is key for the development of abstractive
summarisation systems since summaries written by humans often rephrase
the original content using paraphrases or synonyms or alternative
syntactic constructions.  Recent proposals by \newcite{rush2015neural}
and \newcite{bingel2016text} address this issue.
\newcite{rush2015neural} proposed a neural model for abstractive
compression and summarisation, and \newcite{bingel2016text} proposed a
structured approach to text simplification which jointly predicts
possible compressions and paraphrases.

None of these approaches requires that the input be split into shorter
sentences so that both the corpora used, and the models learned, fail to
adequately account for the various types of specific rephrasings
occurring when a complex sentence is split into several shorter
sentences.

Finally, sentence fusion does induce rephrasing as one sentence is
produced out of several. However, research in that field is still
hampered by the small size of datasets for the task, and the
difficulty of generating one \cite{daume2004generic}. Thus, the
dataset of \newcite{thadani2013supervised} only consists of 1,858
fusion instances of which 873 have two inputs, 569 have three and 416
have four. This is arguably not enough for learning a general
\task\ model.

In sum, while work on sentence rewriting has made some contributions
towards learning to split and/or to rephrase, the interaction between
these two subtasks have never been extensively studied nor are there
any corpora available that would support the development of models
that can both split and rephrase.  In what follows, we introduce such
a benchmark and present some baseline models which provide some interesting 
insights on how to address the \task\ problem.

\section{The \websplit\ Benchmark}
\label{sec:benchmark}



We derive a \task\ dataset from the \webnlg\ corpus presented in  
\newcite{gardent2017creating}.

\subsection{The \webnlg\ Dataset}
\label{subsec:webnlg}

\begin{figure*}[ht]{\footnotesize
\begin{tabular}{ll}\hline
& \webnlg\\\hline
$M_1$ & $\{$ \sem{Birmingham$\mid$leaderName$\mid$John\_Clancy\_(Labour\_politician),}  \sem{John\_Madin$\mid$birthPlace$\mid$Birmingham,}\\
& \sem{103\_Colmore\_Row$\mid$architect$\mid$John\_Madin}$\}$ \\
$T_1^1$ & \nlt{John Clancy is a labour politican who leads Birmingham, where architect John Madin, who designed 103} \\ & \nlt{Colmore Row, was born.}\\
$T_1^2$ & 
\nlt{Labour politician, John Clancy is the leader of Birmingham.}\\
& \nlt{John Madin was born in this city.} \\
& \nlt{He was the architect of 103 Colmore Row.}\\\hline
$M_2$ & $\{$ \sem{Birmingham$\mid$leaderName$\mid$John\_Clancy\_(Labour\_politician)}$\}$\\
$T_2$ & 
\nlt{Labour politician, John Clancy is the leader of Birmingham.}\\\hline
$M_3$ & $\{$ \sem{John\_Madin$\mid$birthPlace$\mid$Birmingham,} \sem{103\_Colmore\_Row$\mid$architect$\mid$John\_Madin}$\}$ \\
$T_3$  &
\nlt{John Madin was born in Birmingham.}\\
& \nlt{He was the architect of 103 Colmore Row.}\\\hline \hline
& \websplit\\\hline
$M_C (=M_1)$ & $\{$ \sem{Birmingham$\mid$leaderName$\mid$John\_Clancy\_(Labour\_politician),} \sem{John\_Madin$\mid$birthPlace$\mid$Birmingham,}\\
& \sem{103\_Colmore\_Row$\mid$architect$\mid$John\_Madin}$\}$ \\
$C (= T_1^1)$ & \nlt{John Clancy is a labour politican who leads Birmingham, where architect John Madin, who designed 103} \\ & \nlt{Colmore Row, was born.}\\
$M_2$ & $\{$ \sem{Birmingham$\mid$leaderName$\mid$John\_Clancy\_(Labour\_politician)}$\}$\\
$T_2$ & 
\nlt{Labour politician, John Clancy is the leader of Birmingham.}\\
$M_3$ & $\{$ \sem{John\_Madin$\mid$birthPlace$\mid$Birmingham,} \sem{103\_Colmore\_Row$\mid$architect$\mid$John\_Madin}$\}$\\
$T_3$  & \nlt{John Madin was born in Birmingham.}\\
& \nlt{He was the architect of 103 Colmore Row.}\\\hline
$M_C (=M_1)$ & $\{$ \sem{Birmingham$\mid$leaderName$\mid$John\_Clancy\_(Labour\_politician),} \sem{John\_Madin$\mid$birthPlace$\mid$Birmingham,}\\
& \sem{103\_Colmore\_Row$\mid$architect$\mid$John\_Madin}$\}$ \\
$C (= T_1^1)$ & \nlt{John Clancy is a labour politican who leads Birmingham, where architect John Madin, who designed 103} \\ & \nlt{Colmore Row, was born.}\\
$M_1$ & $\{$ \sem{Birmingham$\mid$leaderName$\mid$John\_Clancy\_(Labour\_politician),} \sem{John\_Madin$\mid$birthPlace$\mid$Birmingham,}\\
& \sem{103\_Colmore\_Row$\mid$architect$\mid$John\_Madin}$\}$ \\
$T_1^2$ & 
\nlt{Labour politician, John Clancy is the leader of Birmingham.}\\
& \nlt{John Madin was born in this city.} \\
& \nlt{He was the architect of 103 Colmore Row.}\\\hline
\end{tabular}}
\caption{Example entries from the \webnlg\ benchmark and their pairing
  to form entries in the \websplit\ benchmark.}\label{fig:webitem}
\end{figure*}

In the \webnlg\ dataset, each item consists of a set of RDF triples
($M$) and one or more texts ($T_i$) verbalising those triples.

An RDF (Resource Description Format) triple is a triple of the form
\sem{subject$\mid$property$\mid$object} where the subject is a URI
(Uniform Resource Identifier), the property is a binary relation and
the object is either a URI or a literal value such as a string, a date
or a number. In what follows, we refer to the sets of triples
representing the meaning of a text as its meaning representation (MR).
Figure~\ref{fig:webitem} shows three example \webnlg\ items with $M_1
, M_2 , M_3$ the sets of RDF triples representing the meaning of each
item, and $\{T_1^1 , T_1^2\}$, $\{T_2\}$ and $\{T_3\}$ listing
possible verbalisations of these meanings.

The \webnlg\ dataset\footnote{We use a version from February 2017
  given to us by the authors. A more recent version is available here:
  \url{http://talc1.loria.fr/ webnlg/stories/challenge.html}.}
consists of 13,308 MR-Text pairs, 7049 distinct MRs, 1482 RDF entities
and 8 DBpedia categories (Airport, Astronaut, Building, Food,
Monument, SportsTeam, University, WrittenWork). The number of RDF
triples in MRs varies from 1 to 7. The number of distinct RDF tree
shapes in MRs is 60.

\subsection{Creating the \websplit\ Dataset}

To construct the \task\ dataset, we make use of the fact that the
\webnlg\ dataset (i) associates texts with sets of RDF triples and
(ii) contains texts of different lengths and complexity corresponding
to different subsets of RDF triples. The idea is the following. Given
a \webnlg\ MR-Text pair of the form $(M, T)$ where $T$ is a single
complex sentence, we search the \webnlg\ dataset for a set $\{(M_1,
T_1), \ldots, (M_n, T_n)\}$ such that $\{ M_1, \ldots, M_n \}$ is a
partition of $M$ and $\langle T_1, \ldots, T_n \rangle$ forms a text
with more than one sentence. To achieve this, we proceed in three main
steps as follows.



\paragraph{Sentence segmentation} 

We first preprocess all 13,308 distinct verbalisations contained in
the \webnlg\ corpus using the Stanford CoreNLP pipeline
\cite{manning-EtAl:2014:P14-5} to segment each verbalisation $T_i$ into
sentences.

Sentence segmentation allows us to associate each text $T$ in the
\webnlg\ corpus with the number of sentences it contains. This is
needed to identify complex sentences with no split (the input to the
\task\ task) and to know how many sentences are associated with a
given set of RDF triples (e.g., 2 triples may be realised by a single
sentence or by two). As the CoreNLP sentence segmentation often fails
on complex/rare named entities thereby producing unwarranted splits,
we verified the sentence segmentations produced by the CoreNLP
sentence segmentation module for each \webnlg\ verbalisation and
manually corrected the incorrect ones.

\paragraph{Pairing} 

Using the semantic information given by \webnlg\ RDF triples and the
information about the number of sentences present in a \webnlg\ text
produced by the sentence segmentation step, we produce all items of
the form $\langle (M_C , C), \{(M_1 , T_1) \ldots (M_n , T_n) \}
\rangle$ such that:

\begin{itemize}
\item $C$ is a single sentence with semantics $M_C$.
\item $T_1 \ldots T_n $ is a sequence of texts that contains at least
  two sentences.
\item The disjoint union of the semantics $M_{1} \ldots M_{n}$ of the
  texts $T_1 \ldots T_n $ is the same as the semantics $M_C$
  of the complex sentence $C$. That is, $M_C = M_1 \biguplus \ldots
  \biguplus M_n$.
\end{itemize}

This pairing is made easy by the semantic information contained in the
\webnlg\ corpus and includes two subprocesses depending on whether
complex and split sentences come from the same \webnlg\ entry or not.

\textit{Within entries.} 
Given a set of RDF triples $M_C$, a \webnlg\ entry will usually
contain several alternative verbalisations for $M_C$ (e.g.,
$T_1^1$ and $T_1^2$ in Figure~\ref{fig:webitem} are two possible
verbalisations of $M_1$). We first search for entries where one
verbalisation $T_C$ consists of a single sentence and another
verbalisation $T$ contains more than one sentence. For such cases, we
create an entry of the form $\langle (M_C, T_C ), \{(M_C , T) \}
\rangle$ such that, $T_C$ is a single sentence and $T$ is a text
consisting of more than one sentence. The second example item for
\websplit\ in Figure~\ref{fig:webitem} presents this case. It uses
different verbalisations ($T_1^1$ and $T_1^2$) of the same meaning
representation $M_1$ in \webnlg\ to construct a \websplit\ item
associating the complex sentence ($T_1^1$) with a text ($T_1^2$) made of
three short sentences.

\textit{Across entries.}  
Next we create $\langle (M, C ), \{(M_1 , T_1) \ldots (M_n , T_n) \}
\rangle$ entries by searching for all \webnlg\ texts $C$ consisting of
a single sentence. For each such text, we create all possible
partitions of its semantics $M_C$ and for each partition, we search
the \webnlg\ corpus for matching entries i.e., for a set $S$ of
$(M_i,T_i)$ pairs such that (i) the disjoint union of the semantics
$M_i$ in $S$ is equal to $M_C$ and (ii) the resulting set of texts
contains more than one sentence. The first example item for
\websplit\ in Figure~\ref{fig:webitem} is a case in point.  $C (=
T_1^1)$ is the single, complex sentence whose meaning is represented
by the three triples $M$. $\langle T_2 , T_3 \rangle$ is the sequence
of shorter texts $C$ is mapped to. And the semantics $M_2$ and $M_3$
of these two texts forms a partition over $M$.

\paragraph{Ordering.} 
For each item $\langle (M_C, C ), \{(M_1 , T_1) \ldots (M_n , T_n) \}
\rangle$ produced in the preceding step, we determine an order on $T_1
\ldots T_n $ as follows. We observed that the \webnlg\ texts
mostly\footnote{As shown by the examples in Figure~\ref{fig:webitem},
  this is not always the case. We use this constraint as a heuristic
  to determine an ordering on the set of sentences associated with
  each input.}  follow the order in which the RDF triples are
presented. Since this order corresponds to a left-to-right depth-first
traversal of the RDF tree, we use this order to order the sentences in
the $T_i$ texts. 

\subsection{Results}
\label{subsec:benchmarkresults}

By applying the above procedure to the \webnlg\ dataset, we create
1,100,166 pairs of the form $\langle (M_C, T_C), \{(M_1,T_1) \ldots
(M_n , T_n)\} \rangle$ where $T_C$ is a complex sentence and $T_1
\ldots T_n $ is a sequence of texts with semantics $M_1, \ldots M_n $
expressing the same content $M_C$ as $T_C$. 1,945 of these pairs were
of type ``Within entries'' and the rest were of type ``Across
entries''. In total, there are 1,066,115 distinct $\langle T_C, T_1
\ldots T_n \rangle$ pairs with 5,546 distinct complex
sentences. Complex sentences are associated with 192.23 rephrasings in
average (min: 1, max: 76283, median: 16). The number of sentences in
the rephrasings varies between 2 and 7 with an average of 4.99. The
vocabulary size is 3,311.

\section{Problem Formulation}
\label{sec:problemdef}

The \task\ task can be defined as follows. Given a complex sentence
$C$, the aim is to produce a simplified text $T$ consisting of a
sequence of texts $T_1\ldots T_n $ such that $T$ forms a text of at
least two sentences and the meaning of $C$ is preserved in $T$. In
this paper, we proposed to approach this problem in a supervised
setting where we aim to maximise the likelihood of $T$ given $C$ and
model parameters $\theta$: $P(T|C;\theta)$. To exploit the different
levels of information present in the \websplit\ benchmark, we break
the problem in the following ways:

{\footnotesize
\begin{align}
  P(T|C;\theta) &= \sum_{M_C} P(T|C;M_C;\theta) P(M_C|C;\theta) \\ 
  &= P(T|C;M_C;\theta), \mbox{if } M_C \mbox{ is known.} \label{eq:mseq2seq}\\
  &= \sum_{M_{1-n}} 
  \begin{aligned} 
    & P(T|C;M_C;M_{1-n};\theta) \times \\ 
    & P(M_{1-n}|C;M_C;\theta) \label{eq:partgen}
  \end{aligned}
\end{align}
}

where, $M_C$ is the meaning representation of $C$ and $M_{1-n}$ is a
set $\{M_1,\ldots,M_n\}$ which partitions $M_C$.

\section{Split-and-Rephrase Models}
\label{sec:baselines}

In this section, we propose five different models which aim to
maximise $P(T|C;\theta)$ by exploiting different levels of information in
the \websplit\ benchmark.

\subsection{A Probabilistic, Semantic-Based Approach} 
\label{subsec:deepsem}

\newcite{narayan2014hybrid} describes a sentence simplification
approach which combines a probabilistic model for splitting and
deletion with a phrase-based statistical machine translation (SMT) and
a language model for rephrasing (reordering and substituting
words). In particular, the splitting and deletion components exploit
the deep meaning representation (a Discourse Representation Structure,
DRS) of a complex sentence produced by Boxer
\cite{curran2007linguistically}.



Based on this approach, we create a \task\ model (aka \hybridsimpl ) by
(i) including only the splitting and the SMT models (we do not learn
deletion) and (ii) training the model on the \websplit\ corpus.

\subsection{A Basic Sequence-to-Sequence Approach} 
\label{subsec:encoderdecoder}

Sequence-to-sequence models (also referred to as encoder-decoder) have
been successfully applied to various sentence rewriting tasks such as
machine translation \cite{sutskever2011generating,bahdanau2014neural},
abstractive summarisation \cite{rush2015neural} and response
generation \cite{shang2015neural}. They first use a recurrent neural
network (RNN) to convert a source sequence to a dense,
fixed-length vector representation (encoder). They then use another recurrent
network (decoder) to convert that vector to a target sequence. 

We use a three-layered encoder-decoder model with LSTM (Long
Short-Term Memory, \cite{hochreiter1997long}) units for the
\task\ task. Our decoder also uses the local-p attention model with
feed input as in \cite{luong15emnlp}. It has been shown that the local
attention model works better than the standard global attention model
of \newcite{bahdanau2014neural}. We train this model (\seqseq ) to
predict, given a complex sentence, the corresponding sequence of
shorter sentences.

The \seqseq\ model is learned on pairs $\langle C , T \rangle$ of
complex sentences and the corresponding text. It directly optimises
$P(T|C;\theta)$ and does not take advantage of the semantic
information available in the \websplit\ benchmark.

\subsection{A Multi-Source Sequence-to-Sequence Approach} 
\label{subsec:multisource}

\begin{table*}[htbp]
\center
  {\footnotesize
    \begin{tabular}{l|lr}
      Model & Task & Training Size\\\hline 
      \hybridsimpl & Given $C$, predict $T$ & 886,857\\ \hline
      \seqseq & Given $C$, predict $T$ & 886,857 \\ \hline
      \mseqseq & Given $C$ and $M_C$, predict $T$ & 886,866\\ \hline
      \splitmseqseq  & Given $C$ and $M_C$,  predict $M_1 \ldots M_n$ & 13,051\\
      & Given $C$ and $M_i$,  predict $T_i$ & 53,470 \\ \hline
      \splitseqseq & Given $C$ and $M_C$,  predict $M_1 \ldots M_n$ & 13,051\\
      & Given $M_i$,  predict $T_i$ & 53,470 \\ \hline
    \end{tabular}
  }
  \caption{Tasks modelled and training data used by
    \task\ models.}\label{tab:baselines}
\end{table*}

In this model, we learn a multi-source model which takes into account
not only the input complex sentence but also the associated set of RDF
triples available in the \websplit\ dataset.  That is, we maximise
$P(T|C;M_C;\theta)$ (Eqn. \ref{eq:mseq2seq}) and learn a
model to predict, given a complex sentence $C$ and its semantics
$M_C$, a rephrasing of $C$.

As noted by \newcite{gardent2017creating}, the shape of the input may
impact the syntactic structure of the corresponding text. For
instance, an input containing a path $(X|P_1|Y) (Y|P_2|Z)$ equating
the object of a property $P_1$ with the subject of a property $P_2$
may favour a verbalisation containing a subject relative (\nl{x V$_1$
  y who V$_2$ z}). Taking into account not only the sentence $C$ that
needs to be rephrased but also its semantics $M_C$ may therefore help
learning.

We model $P(T|C;M_C;\theta)$ using a multi-source sequence-to-sequence
neural framework (we refer to this model as \mseqseq ). The core idea
comes from \newcite{zoph2016multi} who show that a multi-source model
trained on trilingual translation pairs $((f, g), h)$ outperforms
several strong single source baselines. We explore a similar
``trilingual'' setting where $f$ is a complex sentence ($C$), $g$ is
the corresponding set of RDF triples ($M_C$) and $h$ is the output
rephrasing ($T$).

We encode $C$ and $M_C$ using two separate RNN encoders. To encode
$M_C$ using RNN, we first linearise $M_C$ by doing a depth-first
left-right RDF tree traversal and then tokenise using the Stanford
CoreNLP pipeline \cite{manning-EtAl:2014:P14-5}.  Like in \seqseq, we
model our decoder with the local-p attention model with feed input as
in \cite{luong15emnlp}, but now it looks at both source encoders
simultaneously by creating separate context vector for each
encoder. For a detailed explanation of multi-source encoder-decoders,
we refer the reader to \newcite{zoph2016multi}.

\subsection{Partitioning and Generating} 
\label{subsec:twosteps}
As the name suggests, the \task\ task can be seen as a task which
consists of two subtasks: (i) splitting a complex sentence into
several shorter sentences and (ii) rephrasing the input sentence to
fit the new sentence distribution. We consider an approach which
explicitly models these two steps (Eqn. \ref{eq:partgen}). A first
model $P(M_1, \ldots, M_n|C;M_C;\theta)$ learns to partition a set $M_C$
of RDF triples associated with a complex sentence $C$ into a disjoint
set $\{M_1, \ldots, M_n\}$ of sets of RDF triples. Next, we generate a
rephrasing of $C$ as follows:

\begin{align}
  & P(T|C;M_C;M_1, \ldots, M_n;\theta) \label{eqn:beforeapprox}\\
  & \approx P(T|C;M_1, \ldots, M_n;\theta) \label{eqn:afterapprox} \\
  & = P(T_1,\ldots,T_n|C;M_1, \ldots, M_n;\theta)\\
  & = \prod_i^n P(T_i|C;M_i;\theta)
\end{align}

where, the approximation from Eqn.~\ref{eqn:beforeapprox} to
Eqn.~\ref{eqn:afterapprox} derives from the assumption that the
generation of $T$ is independent of $M_C$ given $(C;M_1, \ldots,
M_n)$. We propose a pipeline model to learn parameters $\theta$. We
first learn to split and then learn to generate from each RDF subset
generated by the split.

\paragraph{Learning to split.} 
For the first step, we learn a probabilistic model which given a set
of RDF triples $M_C$ predicts a partition $M_1 \ldots M_n$ of this
set. For a given $M_C$, it returns the partition $M_1 \ldots M_n$ with
the highest probability $P(M_1, \ldots, M_n|M_C)$.

We learn this split module using items $\langle (M_C, C ), \{(M_1 ,
T_1) \ldots (M_n , T_n) \} \rangle$ in the \websplit\ dataset by
simply computing the probability $P(M_1, \ldots, M_n|M_C)$. To make
our model robust to an unseen $M_C$, we strip off named-entities and
properties from each RDF triple and only keep the tree skeleton of
$M_C$. There are only 60 distinct RDF tree skeletons, 1,183 possible
split patterns and 19.72 split candidates in average for each tree
skeleton, in the \websplit\ dataset.

\paragraph{Learning to rephrase.} 
We proposed two ways to estimate $P(T_i|C;M_i;\theta)$: (i) we learn a
multi-source encoder-decoder model which generates a text $T_i$
given a complex sentence $C$ and a set of RDF triples
$M_i \in M_C$ ; and (ii) we approximate $P(T_i|C;M_i;\theta)$ by
$P(T_i|M_i;\theta)$ and learn a simple sequence-to-sequence model
which, given $M_i$, generates a text $T_i$. Note that
as described earlier, $M_i$'s are linearised and tokenised before we
input them to RNN encoders. We refer to the first model by
\splitmseqseq\ and the second model by \splitseqseq .

\section{Experimental Setup and Results}
\label{sec:setupresults}

This section describes our experimental setup and results. We also
describe the implementation details to facilitate the replication of
our results.

\subsection{Training, Validation and Test sets}
To ensure that complex sentences in validation and test sets are not
seen during training, we split the 5,546 distinct complex sentences in
the \websplit data into three subsets: Training set (4,438, 80\%),
Validation set (554, 10\%) and Test set (554, 10\%).

Table~\ref{tab:baselines} shows, for each of the 5 models, a summary
of the task and the size of the training corpus. For the models that
directly learn to map a complex sentence into a meaning preserving
sequence of at least two sentences ( \hybridsimpl , \seqseq\ and
\mseqseq ), the training set consists of 886,857 $\langle C , T
\rangle$ pairs with $C$ a complex sentence and $T$, the corresponding
text. In contrast, for the pipeline models which first partition the
input and then generate from RDF data (\splitmseqseq and \splitseqseq
), the training corpus for learning to partition consists of 13,051
$\langle M_C , \langle M_1 \ldots M_n \rangle \rangle$ pairs while the
training corpus for learning to generate contains 53,470 $\langle M_i
, T_i \rangle$ pairs.

\subsection{Implementation Details}
For all our neural models, we train RNNs with three-layered LSTM
units, 500 hidden states and a regularisation dropout with probability
0.8. All LSTM parameters were randomly initialised over a uniform
distribution within [-0.05, 0.05]. We trained our models with
stochastic gradient descent with an initial learning rate 0.5. Every
time perplexity on the held out validation set increased since it was
previously checked, then we multiply the current learning rate by
0.5. We performed mini-batch training with a batch size of 64
sentences for \seqseq\ and \mseqseq, and 32 for \splitseqseq\ and
\splitmseqseq. As the vocabulary size of the \websplit\ data is small,
we train both encoder and decoder with full vocabulary. We randomly
initialise word embeddings in the beginning and let the model train
them during training. We train our models for 20 epochs and keep the
best model on the held out set for the testing purposes. We used the
system of \newcite{zoph2016multi} to train both simple
sequence-to-sequence and multi-source sequence-to-sequence
models\footnote{We used the code available at
  \url{https://github.com/isi-nlp/Zoph_RNN}.}, and the system of
\newcite{narayan2014hybrid} to train our
\hybridsimpl\ model.\footnote{We used the code available at
  \url{https://github.com/shashiongithub/Sentence-Simplification-ACL14}.}

\begin{table}[htbp]{\footnotesize
\begin{tabular}{l | r r r }
Model & BLEU & \#S/C & \#Tokens/S \\\hline
\textsc{Source} & 55.67 & 1.0 & 21.11  \\
\hybridsimpl & 39.97 & 1.26 & 17.55 \\
\seqseq & 48.92 & 2.51 & 10.32  \\
\mseqseq & 42.18 & 2.53 & 10.69  \\
\splitmseqseq & 77.27 & \textbf{2.84} & 11.63 \\ 
\splitseqseq & \textbf{78.77} & \textbf{2.84} & \textbf{9.28} \\
\hline
\end{tabular}}
\caption{Average BLEU scores for rephrasings, average number of sentences in the output texts (\#S/C) and average number of tokens per output sentences (\#Tokens/S). \textsc{Source} are the complex sentences from the \websplit\ corpus. }\label{tab:results}
\vspace{-0.3cm}
\end{table}

\begin{table*}[htbp]{\footnotesize
\begin{tabular}{ll}
\textsc{Source} & Alan Shepard was born in New Hampshire and he served as the Chief of the Astronaut Office .\\\hline
\hybridsimpl & Alan Shepard was born in New Hampshire and he served as of the the chief astronaut office .\\
\seqseq & Alan Shepard 's occupation was a test pilot . Alan Shepard was born in New Hampshire . \\& Alan Shepard was born on Nov 18 , 1923 .\\
\mseqseq &  Alan Shepard served as a test pilot . Alan Shepard 's birth place was New Hampshire .\\
\splitmseqseq &  Alan Shepard served as the Chief of the Astronaut Office . Alan Shepard was born in New \\ 
& Hampshire .\\
\splitseqseq & Alan Shepard served as the Chief of the Astronaut Office . Alan Shepard 's birth place was\\ & New Hampshire .\\ \hline
\end{tabular}}
\caption{Example outputs from different models.}\label{tab:outputs}
\vspace{-0.3cm}
\end{table*}
\subsection{Results}
\label{subsec:results}

We evaluate all models using multi-reference BLEU-4 scores
\cite{Papineni:2002} based on all the rephrasings present in the
\task\ corpus for each complex input sentence.\footnote{We used
  \url{https://github.com/moses-smt/mosesdecoder/blob/master/scripts/generic/multi-bleu.perl}
  to estimate BLEU scores against multiple references.}  As BLEU is a
metric for $n$-grams precision estimation, it is not an optimal metric
for the \task\ task (sentences even without any split could have a
high BLEU score). We therefore also report on the average number of
output simple sentences per complex sentence and the average number of
output words per output simple sentence. The first one measures the
ability of a system to split a complex sentence into multiple simple
sentences and the second one measures the ability of producing smaller
simple sentences.

Table~\ref{tab:results} shows the results. The high BLEU score for
complex sentences (\textsc{Source}) from the \websplit\ corpus shows
that using BLEU is not sufficient to evaluate splitting and
rephrasing. Because the short sentences have many n-grams in common
with the source, the BLEU score for complex sentences is high but the
texts are made of a single sentence and the average sentence length is
high. \hybridsimpl\ performs poorly -- we conjecture that this is
linked to a decrease in semantic parsing quality (DRSs) resulting from
complex named entities not being adequately recognised. The simple
sequence-to-sequence model does not perform very well neither does the
multi-source model trained on both complex sentences and their
semantics. Typically, these two models often produce non-meaning
preserving outputs (see example in Table~\ref{tab:outputs}) for input
of longer length. In contrast, the two partition-and-generate models
outperform all other models by a wide margin. This suggests that the
ability to split is key to a good rephrasing: by first splitting the
input semantics into smaller chunks, the two partition-and-generate
models permit reducing a complex task (generating a sequence of
sentences from a single complex sentence) to a series of simpler tasks
(generating a short sentence from a semantic input).

Unlike in neural machine translation setting, multi-source models in
our setting do not perform very well. \seqseq\ and
\splitseqseq\ outperform \mseqseq\ and \splitmseqseq\ respectively,
despite  using less input information than their counterparts.
The multi-source models used in machine translation have as a
multi-source, two translations of the same content
\cite{zoph2016multi}. In our approach, the multi-source is a complex
sentence and a set of RDF triples, e.g., $(C;M_C)$ for \mseqseq\ and
$(C;M_i)$ for \splitmseqseq. We conjecture that the poor performance
of multi-source models in our case is due either to the relatively
small size of the training data or to a stronger mismatch between RDF
and complex sentence than between two translations.

Table~\ref{tab:outputs} shows an example output for all 5 systems
highlighting the main differences.  \hybridsimpl's output mostly
reuses the input words suggesting that the SMT system doing the
rewriting has limited impact. Both the \seqseq\ and the
\mseqseq\ models ``hallucinate'' new information (\nl{served as a test
  pilot}, \nl{born on Nov 18, 1983}). In contrast, the
partition-and-generate models correctly render the meaning of the
input sentence (\textsc{Source}), perform interesting rephrasings
(\nl{X was born in Y} $\rightarrow$ \nl{X's birth place was Y}) and
split the input sentence into two.

\section{Conclusion}
\label{sec:conclusion}

We have proposed a new sentence simplification task which we call
``\task ''.  We have constructed a new corpus for this task which is
built from readily-available data used for NLG (Natural Language
Generation) evaluation. Initial experiments indicate that the ability
to split is a key factor in generating fluent and meaning preserving
rephrasings because it permits reducing a complex generation task
(generating a text consisting of at least two sentences) to a series
of simpler tasks (generating short sentences). In future work, it
would be interesting to see whether and if so how, sentence splitting
can be learned in the absence of explicit semantic information in the
input.

Another direction for future work concerns the exploitation of the
extended WebNLG corpus. While the results presented in this paper use
a version of the WebNLG corpus consisting of 13,308 MR-Text pairs,
7049 distinct MRs and 8 DBpedia categories, the current WebNLG corpus
encompasses 43,056 MR-Text pairs, 16,138 distinct MRs and 15 DBpedia
categories. We plan to exploit this extended corpus to make available
a correspondingly extended \websplit\ corpus, to learn optimised
\task\ models and to explore sentence fusion (converting a sequence
of sentences into a single complex sentence).

\section*{Acknowledgements}
We thank Bonnie Webber and Annie Louis for early discussions on the
ideas presented in the paper. We thank Rico Sennrich for directing us
to multi-source NMT models. This work greatly benefited from
discussions with the members of the Edinburgh NLP group. We also thank
the three anonymous reviewers for their comments to improve the paper.
The research presented in this paper was partially supported by the
H2020 project SUMMA (under grant agreement 688139) and the French
National Research Agency within the framework of the WebNLG Project
(ANR-14-CE24-0033).

\bibliography{simplification}
\bibliographystyle{emnlp_natbib}
\end{document}